\documentclass[sigconf]{acmart}

\AtBeginDocument{%
  \providecommand\BibTeX{{%
    \normalfont B\kern-0.5em{\scshape i\kern-0.25em b}\kern-0.8em\TeX}}}

\copyrightyear{2023}
\acmYear{2023}
\setcopyright{rightsretained}
\acmConference[C\&C'23]{Creativity and Cognition}{June 19--21, 2023}{Virtual Event, USA}
\acmBooktitle{Creativity and Cognition (C\&C23), June 19--21, 2023, Virtual Event, USA}\acmDOI{10.1145/3591196.3593516}
\acmISBN{979-8-4007-0180-1/23/06}

\usepackage{multirow}
\usepackage{subcaption}
\usepackage{caption}
\usepackage{chngcntr}

\begin{document}

%%
%% The "title" command has an optional parameter,
%% allowing the author to define a "short title" to be used in page headers.
\title[Large Language Models' Capacity for Augmenting Cross-Domain Analogical Creativity]{Fluid Transformers and Creative Analogies: Exploring Large Language Models' Capacity for Augmenting Cross-Domain Analogical Creativity}

%%
%% The "author" command and its associated commands are used to define
%% the authors and their affiliations.
%% Of note is the shared affiliation of the first two authors, and the
%% "authornote" and "authornotemark" commands
%% used to denote shared contribution to the research.
\author{Zijian Ding}
\affiliation{%
  \institution{College of Information Studies, University of Maryland}
  \country{USA}}
  
\author{Arvind Srinivasan}
\affiliation{%
  \institution{College of Information Studies, University of Maryland}
  \country{USA}}

\author{Stephen MacNeil}
\affiliation{%
  \institution{Computer and Information Sciences, Temple University}
  \country{USA}}
  
\author{Joel Chan}
\affiliation{%
  \institution{College of Information Studies, University of Maryland}
  \country{USA}}
  
% \author{Anonymous}
% \affiliation{%
%   \institution{Anonymous}}
  
% \author{Anonymous}
% \affiliation{%
%   \institution{Anonymous}}

%%
%% By default, the full list of authors will be used in the page
%% headers. Often, this list is too long, and will overlap
%% other information printed in the page headers. This command allows
%% the author to define a more concise list
%% of authors' names for this purpose.
\renewcommand{\shortauthors}{Ding, et al.}

%%
%% The abstract is a short summary of the work to be presented in the
%% article.
\begin{abstract}
 Cross-domain analogical reasoning is a core creative ability that can be challenging for humans. Recent work has shown some proofs-of-concept of Large language Models’ (LLMs) ability to generate cross-domain analogies. However, the reliability and potential usefulness of this capacity for augmenting human creative work has received little systematic exploration. In this paper, we systematically explore LLMs capacity to augment cross-domain analogical reasoning.  Across three studies, we found: 1) LLM-generated cross-domain analogies were frequently judged as helpful in the context of a problem reformulation task (median 4 out of 5 helpfulness rating), and frequently ($\sim$80\% of cases) led to observable changes in problem formulations, and 2) there was an upper bound of $\sim$25\% of outputs being rated as potentially harmful, with a majority due to potentially upsetting content, rather than biased or toxic content. These results demonstrate the potential utility --- and risks --- of LLMs for augmenting cross-domain analogical creativity.  %Results show that LLMs are able to consistently generate useful analogies for creative tasks with a harmfulness rate of 26\%.
\end{abstract}

%%
%% The code below is generated by the tool at http://dl.acm.org/ccs.cfm.
%% Please copy and paste the code instead of the example below.
%%


% \ccsdesc[500]{Human-centered computing~Empirical studies in HCI}
% \ccsdesc[500]{Human-centered computing~Auditory feedback}
% \ccsdesc[300]{Human-centered computing~Empirical studies in interaction design}
% \ccsdesc[500]{Human-centered computing~Sound-based input / output}

%%
%% Keywords. The author(s) should pick words that accurately describe
%% the work being presented. Separate the keywords with commas.

\maketitle

\def \RQO {\textbf{RQ1}: ?}

\def \RQT {\textbf{RQ2}: ?}

\section{Introduction}
\label{sec:intro}

Large language models (LLMs) such as GPT-3 are attracting attention in the creativity research community for their potential to augment creative work by generating tailored design materials and prototypes. For example, researchers have explored the potential of LLMs to replicate or assist creative writing tasks such as metaphor generation \cite{chakrabartyMERMAIDMetaphorGeneration2021,geroMetaphoriaAlgorithmicCompanion2019}, science writing \cite{geroSparksInspirationScience2022} and storytelling \cite{yuanWordcraftStoryWriting2022a, roemmeleWritingStoriesHelp2016, singhWhereHideStolen2022, berkovskyHowNovelistsUse2018, freiknechtProceduralGenerationInteractive2020, osoneBunChoAISupported2021, clarkCreativeWritingMachine2018, cao2023scaffolding}. %, and analogical reasoning \cite{webbEmergentAnalogicalReasoning2022}. 
This research thread parallels explorations in industry, such as the ``AI Dungeon" startup's use of GPT-3 to assist with world-building in Dungeons and Dragons \cite{teamWorldCreationAnalogy2020}, prototype systems for AI-assisted content creation\footnote{https://www.regie.ai/, https://www.copy.ai/, https://writer.com/} and writing assistance\footnote{https://www.compose.ai/}, as well as other applications outside of creative writing, such as Q\&A conversation\footnote{https://chat.openai.com/chat} and code generation\footnote{https://github.com/features/copilot}. %These models' capacity for in-context learning (e.g., using prompt programming paradigms) may also open up a new range of interactive systems that better manage the gap between technical knowledge (to create models) and domain knowledge (task understanding, evaluating performance in a realistic way). 
Here, we are particularly interested in probing the potential of LLMs to augment \textbf{cross-domain analogical creativity}. 

Cross-domain analogical reasoning is the core cognitive ability to perceive and reason about deep structural similarities between situations that may differ on many surface details \cite{holyoakMentalLeapsAnalogy1996,gentnerStructureMappingTheoretical1983,hofstadterFluidConceptsCreative1995}; for example, using analogy, people can recognize a strong similarity between the solar system and the atom in terms of relational similarity (both involve a central mass --- sun, nucleus --- orbited by smaller bodies --- planets, electrons), ignoring other surface dissimilarities such as their relative size and color. Cross-domain analogy is a frequent source of creative breakthroughs, enabling creators to develop powerful new concepts \cite{gentnerAnalogyScientificDiscovery2002,holyoakMentalLeapsAnalogy1996,hargadonTechnologyBrokeringInnovation1997,dahlInfluenceValueAnalogical2002,chanBenefitsPitfallsAnalogies2011} or reformulations of their creative problems \cite{dorstFrameInnovationCreate2015, helmsCompoundAnalogicalDesign2008, okadaAnalogicalModificationCreation2009, bhattaInnovationAnalogicalDesign1994,graceInterpretationdrivenMappingFramework2015}. As an illustrative example, Dorst \cite{dorstFrameInnovationCreate2015} reported a case study of how designers used an analogy to a music festival to (re)frame the problem of night violence in King's Cross from a crime prevention / reduction problem with a new entertainment perspective.

We are motivated by the potential of LLMs to augment creative analogy-making because cross-domain analogies can be hard for humans to retrieve in the creative process. Human retrieval is highly sensitive to surface similarities, favoring ``near", or within-domain, analogs that share attributes of an object over ``far", or structurally similar analogs from different domains that primarily share relations to the object \cite{holyoakMentalLeapsAnalogy1996, gentnerAnalogicalRemindingGood1985, gickSchemaInductionAnalogical1983, gentnerRolesSimilarityTransfer1993}. This can lead to creators failing to retrieve relevant analogs from other domains because they are fixated on surface features of their source problem \cite{linseyModalityRepresentationAnalogy2008, linseyRepresentingAnalogiesIncreasing2006}. For example, people trying to solve Duncker's \cite{dunckerProblemsolving1945} radiation problem ---- how to remove a cancerous tumor using radiation without damaging surrounding healthy tissue --- are much more likely to retrieve analogs involving cancer or radiation than an analog of an army splitting up to attack a target. The ability of LLMs to dynamically generate elaborate text tuned to specific settings presents an opportunity to complement existing structured methods of supporting creative analogy-making \cite{gordonSynecticsDevelopmentCreative1961}, such as the WordTree method \cite{linseyWordTreesMethodDesignbyanalogy2008} or the TRIZ method \cite{savranskyEngineeringCreativityIntroduction2000}, which are manual and effortful,  and search-based systems that may require expensive, specialized pre-processing of databases of potential analogs \cite{fuMeaningFarImpact2013,kangAugmentingScientificCreativity2022,hopeAcceleratingInnovationAnalogy2017,palani2021active,chanSOLVENTMixedInitiative2018,vattamDANEFosteringCreativity2011, palani2021conotate}. 

Past work suggests that LLMs may be able to generate analogies that resemble human-generated analogies. For example, there are substantial systematic investigations of LLM performance on standard analogy completion word problems such as the SAT four-term analogy problems; an example of this task is to generate a term that answers a question like ``man is to woman as king is to ...?" (where a correct answer is ``queen") \cite{brownLanguageModelsAre2020,webbEmergentAnalogicalReasoning2022,bhavyaAnalogyGenerationPrompting2022}. More recently, some researchers have shared proof-of-concept demonstrations of LLMs' ability to generate longer natural-language analogies \cite{teamWorldCreationAnalogy2020}, explanations of analogical mappings \cite{webbEmergentAnalogicalReasoning2022,bhavyaAnalogyGenerationPrompting2022}, or analogy-inspired concepts for creative problems \cite{zhuGenerativePreTrainedTransformer2022,zhuBiologicallyInspiredDesign2023,webbEmergentAnalogicalReasoning2022}.
% However, because we want to understand LLMs ability to augment --- rather than automate --- humans' creative processes, we are not primarily interested in the degree to which LLMs exhibit human-like performance on analogy (which is the primary focus of existing NLP research on this topic). In particular, benchmark measures based on correctness are helpful for understanding whether/how LLMs can recognize or generate analogies, but 

However, with some exceptions \cite{zhuBiologicallyInspiredDesign2023}, prior work lacks systematic and direct investigation of how these analogies might be \textit{useful} in the creative process. We want to answer questions like, can people use LLM-generated analogies, even if they may be ``incorrect" (e.g., missing key relational mappings between a source problem and target analogy), to inspire problem reformulation or ideation? If so, to what extent does this happen? Can we predict in advance which LLM-generated analogies might turn out to be useful? Given concerns about potential bias/harm in LLMs \cite{benderDangersStochasticParrots2021, bommasaniOpportunitiesRisksFoundation2021}, how might LLM-generated analogies' potential usefulness for augmented creativity trade off with potential harmfulness or toxicity in LLM outputs?

These questions cannot be satisfactorily answered in a traditional NLP-oriented benchmark paradigm. For example it is common to compare outputs with a ``correct" reference output, as in the analogy completion word problems; this is insufficient to capture the potential usefulness of analogical inspirations that may only emerge from usage, or run through the generating and iterating on "bad" ideas \cite{dixWhyBadIdeas2006,gruberDarwinManPsychological1974,chanImportanceIterationCreative2015,kneelandExploringUnchartedTerritory2020}. Crowdsourced human evaluations of LLM outputs may also index only surface-level linguistic coherence vs. more substantive dimensions of quality without more specific (task-specific) instructions: for example, Clark et al \cite{clarkAllThatHuman2021} reported that crowd workers mostly relied on form vs. content heuristics to make their judgments about human-likeness of LLM-generated text; more specific instructions or task framings --- such as contextualizing ratings within a creative task --- may be necessary to move beyond surface judgments.

In this paper, we directly investigate the potential use of LLM-generated analogies in the creative task of design problem reformulation. %on problem (re)formulation as the context of creative use for LLM-generated analogies. %given its importance to the creative process, and prior research on how 
First, following a \textit{prompt-based learning} paradigm \cite{liuPretrainPromptPredict2023} (also called in-context learning in machine learning research \cite{brownLanguageModelsAre2020}), we crafted a structured natural-language instruction prompt for generating cross-domain analogies (including, in some cases, one or more examples of cross-domain analogies). We used these prompts to generate 480 cross-domain analogies across six design problems, and systematically explored human judgments of predicted usefulness of LLM-generated analogies (i.e., not in the context of using analogies in a creative task), along with duplicate rate and semantic distance of analogies from the design problems (\textbf{prompt engineering explorations}; Section \ref{sec:study1}). Second, we collected user ratings of helpfulness of a subset of these analogies in the context of a problem reformulation task, along with content analysis of user reformulation behaviors with the analogies (\textbf{Study 1}; Section \ref{sec:study2}). Third and finally, we manually analyzed the nature and rates of potentially harmful outputs in the analogy generations (\textbf{Study 2}; Section \ref{sec:study3}). %and study where it does well and where it struggles. We lean heavily on extrinsic, use-time evaluation of the models' outputs, given the inherent uncertainty of evaluating creative intermediate and final products \cite{sosaAccretionTheoryIdeation2019}. 

Our primary findings were as follows:
\begin{enumerate}
    \item \textbf{Prompt engineering explorations:} We were able to construct one- or few-shot prompts that yielded cross-domain analogies where ~70\% of outputs were both unique and judged by researcher to be potentially useful for creative problem reformulation.
    \item \textbf{Study 1:} A majority of LLM-generated analogies were rated as helpful for individuals during a creative problem reformulation task, primarily spurring new considerations for the design problems, but also encouraging shifts in problem perspectives and redefining key elements in the original problem statements. Notably, there was no correlation between \textit{a priori} judgments of potential usefulness from prompt engineering explorations and the use-time ratings of analogy helpfulness here. 
    \item \textbf{Study 2:} There was some evidence of harmful or biased/toxic outputs in generated analogies (upper bound of 25\% of outputs as screened by human raters), though the clear majority of potentially harmful outputs ($\sim$80\%) were describing situations that could be conservatively considered upsetting under some circumstances (e.g., describing situations of poverty or difficult childbirth), rather than biased or toxic.
\end{enumerate} 

Overall, our findings suggest that LLM-generated analogies hold potential as a creativity support tool for cross-domain analogical problem reformulation, and extend previous demonstrations of LLMs' capabilities with analogical reasoning for creative settings. %The lack of correlation between the \textit{a priori} usefulness judgments in prompt engineering explorations and the in-use helpfulness ratings and reformulation behaviors in Study 1 also underscore the importance of in-use evaluations of LLM-generated analogies for creative applications, and going beyond simple benchmark-based measures of output ``quality" \cite{leeEvaluatingHumanLanguageModel2022}. 
To facilitate further analyses by the community, we also share the code used to generate the datasets in prompt engineering explorations (along with the human judgment data), and raw and coded participant responses from Study 1. We hope that the rich descriptive data provided in this paper can help researchers of creativity support tools understand how to effectively leverage LLMs to augment cross-domain analogical creativity.

\section{Related Work}
% Our work primarily extends prior research on large language models and analogical reasoning. 

% \subsection{Large language models and analogical reasoning}

The field of natural language processing (NLP) has studied analogies due to their common use in language and their importance for understanding semantic relationships between words and phrases. For example, classic work in NLP focuses on the concept of word embeddings where words like king and queen are statistically likely to be used in the same contexts~\cite{mikolovEfficientEstimationWord2013}.
% {\color{red}Tracing back to at least \cite{mikolovEfficientEstimationWord2013}'s demonstration of the word2vec word embedding model's ability to solve four-term analogies like ("man is to woman as king is to <ans:queen>") using vector-offset arithmetic, a robust body of research has suggested that unsupervised or self-supervised deep learning models of language are able to capture analogical relations to some degree.} 
Careful analyses of pre-transformer architectures, such as word2vec \cite{mikolovEfficientEstimationWord2013}, demonstrated that the surprisingly high accuracy (on the order of 60\%) on analogy word problems (e.g., ``man is to woman as king is to <ans:queen>") hid large variations in accuracy across subtypes of analogy word problems: for example, \cite{chenEvaluatingVectorspaceModels2017} observed consistently higher accuracy on syntactic analogies (e.g., based on morphological transformations like ``big is bigger as small is to <ans:smaller>") compared to semantic analogies based on causal relations \cite{chenEvaluatingVectorspaceModels2017}. Some researchers were nevertheless able to leverage these models as base layers or inputs to NLP pipelines that were able to do analogical retrieval in complex natural language datasets such as crowdsourced ideas \cite{hopeAcceleratingInnovationAnalogy2017}, and research papers \cite{chanSOLVENTMixedInitiative2018}. 
More recent transformer-based models, such as GPT-3, have shown improved performance on this task. For instance, the largest 175B parameter version of GPT-3 (codenamed \textit{davinci}) achieved comparable performance to humans ($\sim$65\% accuracy) on a set of SAT four-term analogy problems, which are considered more challenging than the analogy word problems studied in previous work \cite{brownLanguageModelsAre2020}. There is also some evidence that more recent models that add alignment training procedures \cite{ouyangTrainingLanguageModels2022}, such as GPT-3's text-davinci-003, can match or exceed human benchmark performance on these tasks \cite{webbEmergentAnalogicalReasoning2022,bhavyaAnalogyGenerationPrompting2022, ding2023mapping}.

More importantly for our current purposes, researchers and practitioners have begun to produce proof-of-concept demonstrations of LLMs' ability to generate more complex natural language analogies. For instance, Zhu et al \cite{zhuGenerativePreTrainedTransformer2022} showed examples of using GPT-3's earlier davinci models to generate analogous design concepts when prompted with analogies between a source problem and real-world design. Similarly, Webb et al \cite{webbEmergentAnalogicalReasoning2022} replicated the classic analogical problem solving paradigm from Gick and Holyoak \cite{gickAnalogicalProblemSolving1980} with the text-davinci-003 version of GPT-3, showing that the model was able to generate a convergence solution for Duncker's radiation problem when prompted with the analogous generals story, as well as describing the analogical mapping between the problem and the analogy (though the model failed to generate a plausible analogous solution for a different physics-based problem, despite successfully describing the analogical mapping between the problem and its analogy). In a slightly different paradigm, but also with an aligned version of GPT, Bhavya et al \cite{bhavyaAnalogyGenerationPrompting2022} showed that $\sim$60\% of InstructGPT \cite{ouyangTrainingLanguageModels2022}-generated analogical explanations for scientific concepts were rated by crowd workers as containing a meaningful analogy, a rate comparable to a dataset of human-generated analogical explanations. On the industry/practitioner side, the prominent AI Dungeon startup published a blog post that described its experiments using GPT-3 to generate descriptions of fantasy worlds by analogy \cite{teamWorldCreationAnalogy2020}. In contrast to the simpler analogy word problems, these more complex analogy-generation capabilities have received little systematic evaluation in the context of creative tasks. One exception is a recent study by Zhu et al \cite{zhuBiologicallyInspiredDesign2023}, who followed up on their previous proof-of-concept \cite{zhuGenerativePreTrainedTransformer2022} by integrating 10 analogically generated design concepts into a design team's brainstorming process, and obtained ratings of the novelty and feasibility of the concepts: the analogical concepts generally received high novelty scores but low feasibility scores.

Overall, existing research suggests that LLM-generated analogies might frequently include sufficient cross-domain analogical mappings to inspire creative problem reformulation and ideation. In this study, we seek to extend the predominantly informal, proof-of-concept demonstrations in prior work with systematic, direct evaluation of LLM-generated analogies in the context of a creative task.

\section{Prompt Engineering Explorations: Analogy Generation Quality as a Function of Prompt-Based Learning Design}
\label{sec:study1}
\begin{figure*}
    \centering \includegraphics[width=0.7\textwidth]{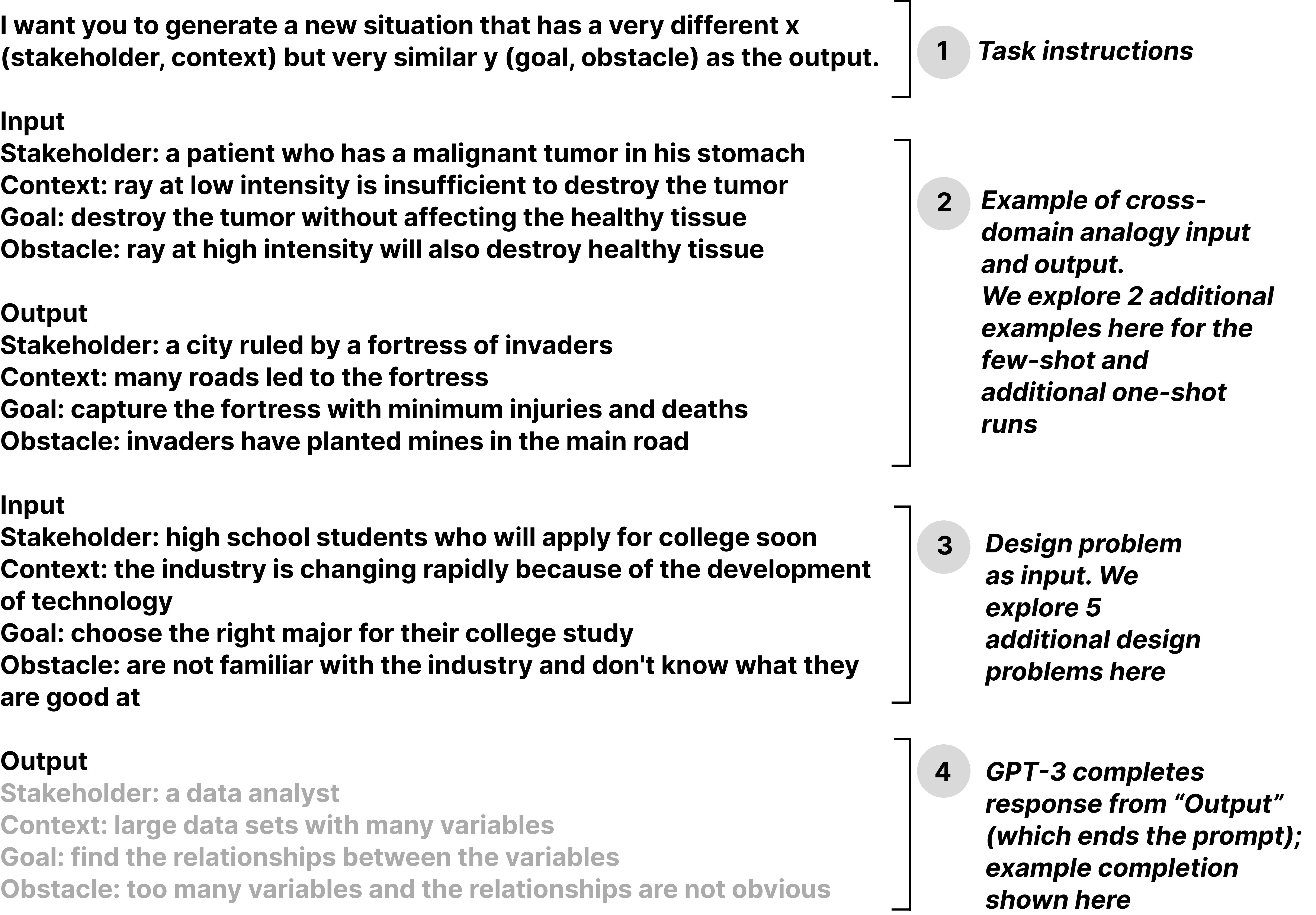}
    \caption{Our prompt-based learning prompts for generating cross-domain analogies.}
\label{fig:prompt}
    \Description{TODO}
\end{figure*}

\begin{figure*}
    \centering \includegraphics[width=1\textwidth]{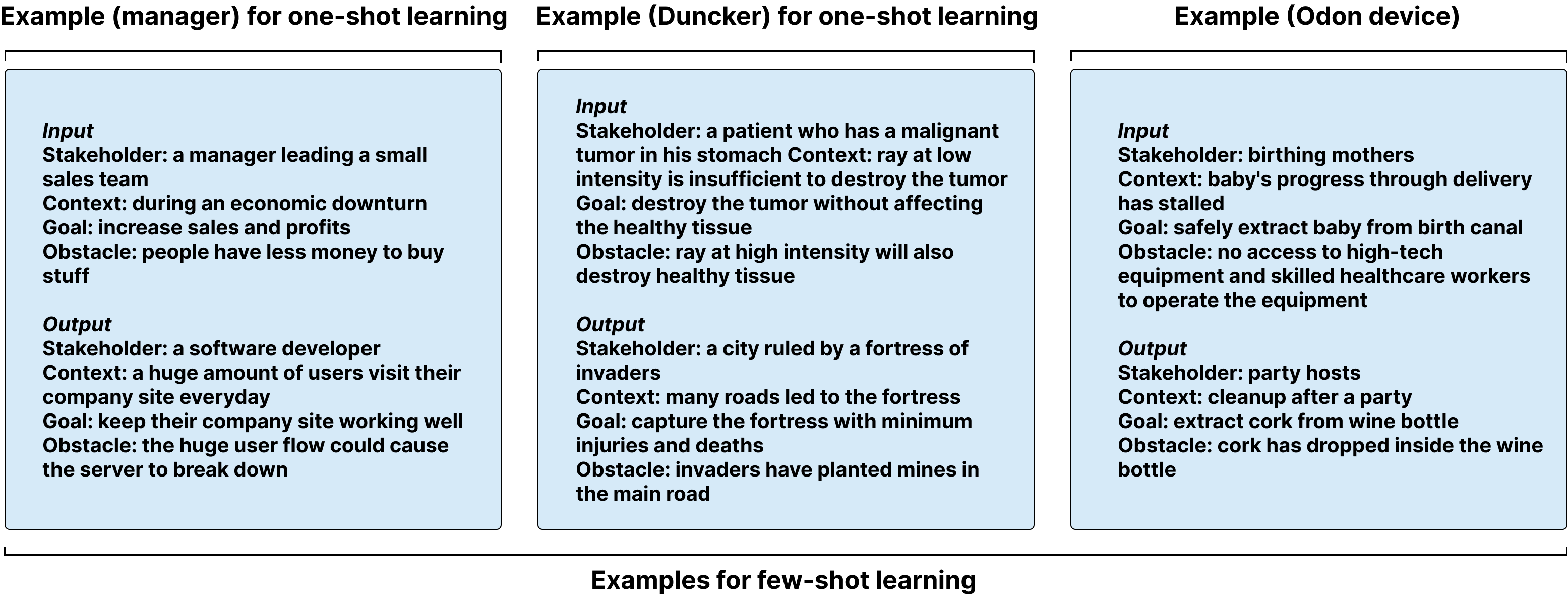}
    \caption{Problem-analogy pairs used as examples in prompts for our one-shot and few-shot learning paradigm runs.}
\label{fig:analogy}
    \Description{TODO}
\end{figure*}

\begin{figure*}
\centering \includegraphics[width=1\textwidth]{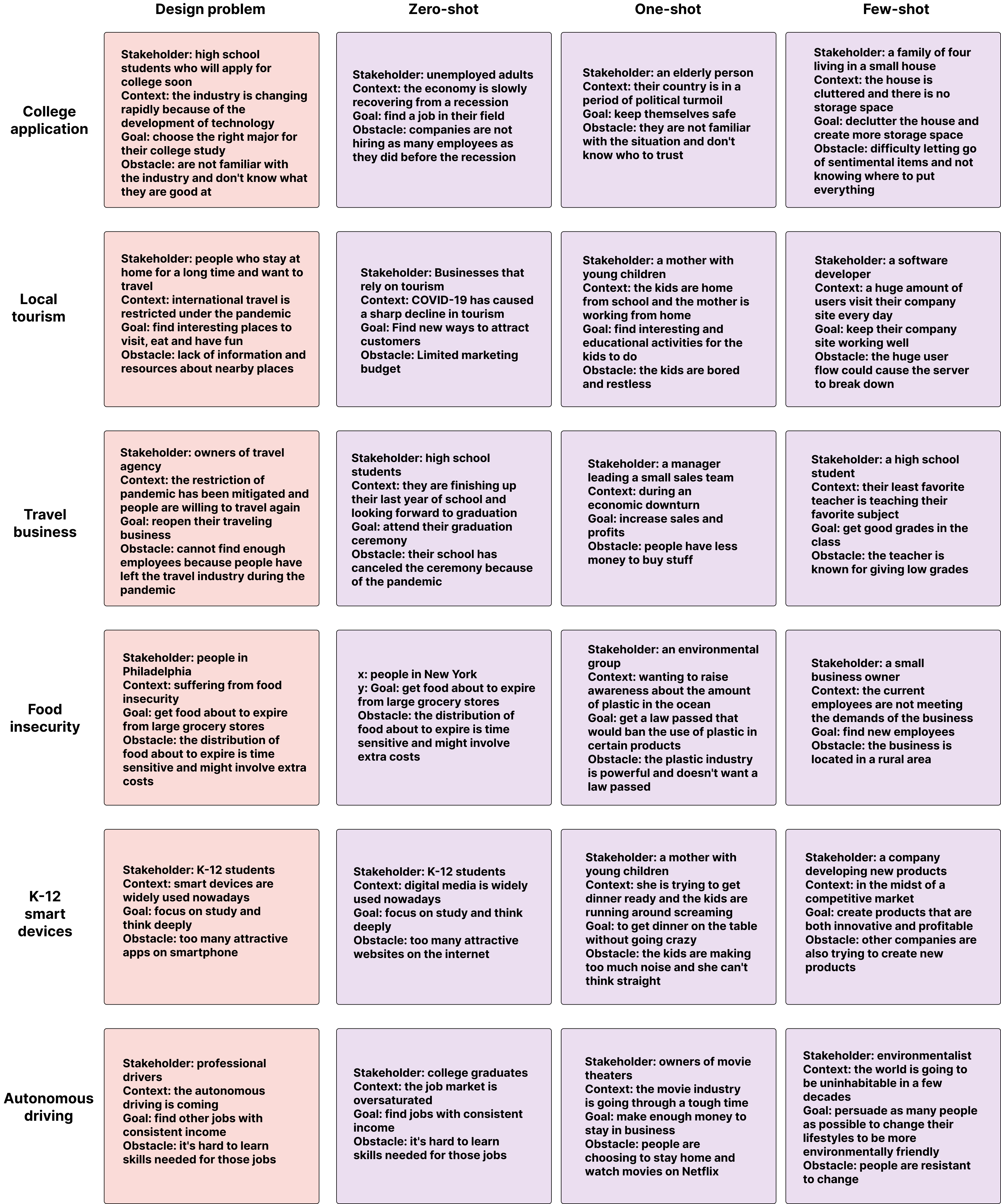}
\caption{Six design problems used for Study 1 and corresponding examples of GPT-3 generated analogies with zero-shot, one-shot, and few-shot paradigms.}
\label{fig:6-problems}
\Description{TODO}
\end{figure*}

\begin{figure*}
    \centering \includegraphics[width=1\textwidth]{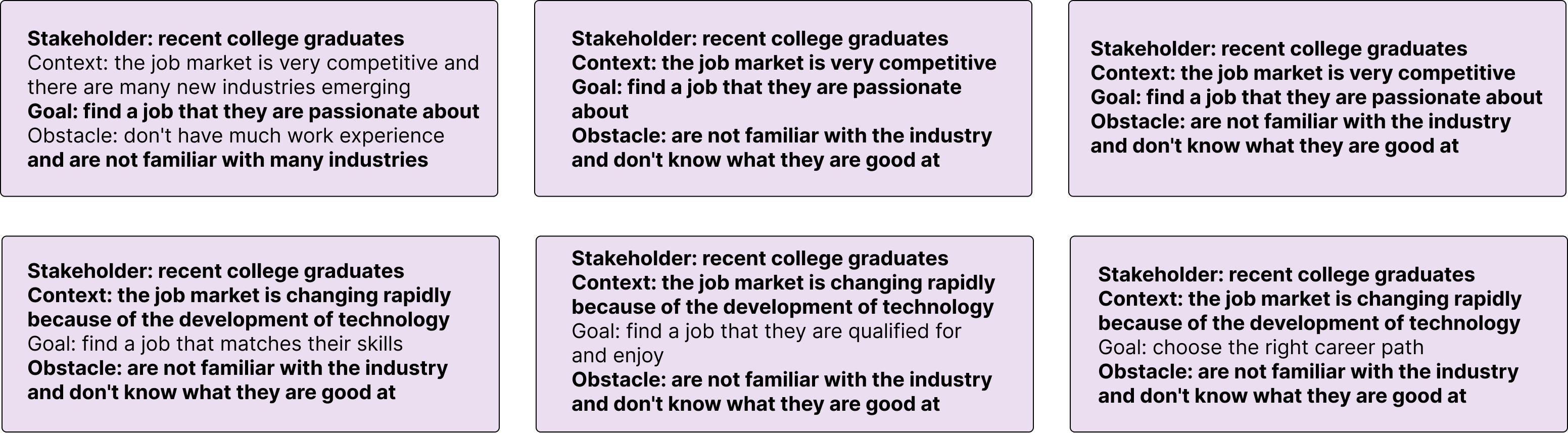}
    \caption{Examples of GPT-3 generated analogies that were judged as duplicates for the same design problem ``choose the right major for their college study'' with zero-shot learning. These duplicate analogies shared very similar concepts/aspects with substantial verbatim overlap in terms (bolded in the figure).}
\label{fig:dup-analogy}
    \Description{TODO}
\end{figure*}

To develop analogies for this study, we followed a \textit{prompt-based learning} paradigm \cite{liuPretrainPromptPredict2023} (also called in-context learning in machine learning research \cite{brownLanguageModelsAre2020}): rather than fine-tuning GPT-3 for analogy generation (e.g., in a supervised learning paradigm), we ``prompted" an LLM by providing a set of natural language instructions for a task of generating cross-domain analogies, and took the LLM's generated text completion as the output. We iteratively improved our prompt through prompt engineering \cite{reynoldsPromptProgrammingLarge2021a}, which involves crafting a prompt for a LLM, issuing the prompt, and evaluating the response from an LLM. %This approach is applied iteratively with the goal of identifying an optimal prompt. 
% For example, an earlier version of our prompt was tried to abstract a single problem statement, an important step in analogy generation \cite{kitturScalingAnalogicalInnovation2019}: "Given a problem statement, generate an abstracted reformulation of that problem statement."

We supplemented this iterative process with a systematic evaluation of the best-performing prompt from the prompt engineering process, exploring zero-, one-, and few-shot learning variants of the prompt. The results from this systematic evaluation are reported in this section.  
%Specifically, we explored variations across zero-shot to few-shot paradigms.

% \begin{figure}
%     \centering \includegraphics[width=0.8\textwidth]{figures/Analogy-example.png}
%     \caption{Example of prompt-output pair (input design problem and GPT-3 generated output) with one-shot learning.}
% \label{fig:gpt3-analogy}
%     \Description{TODO}
% \end{figure}

\subsection{Task setup and prompt-based learning design}

% As a first step, we did the most simple thing of asking the language model to do the whole task, to study where it does well and where it struggles. From here, we will start to decompose the task more.
We conducted our experiments with OpenAI's GPT-3 API, with the text-davinci-002 model, with temperature = 1 for the largest output variety and token count = 400 to accommodate the output length. Because we are interested in exploring how LLM-generated analogies could be integrated into problem formulation, we experimented with a prompt design where we provided a structure of the input problem and the output analogy by dividing a design problem into four components—stakeholder, context, goal and obstacle led to higher quality outputs. This was inspired by previous work on problem formulation in design \cite{macneilFramingCreativeWork2021}, and in our initial prompt engineering iterations, we informally observed that this enabled us to balance the distance (stakeholder and context are different) and usefulness (goal and obstacle are similar) of the generated analogy much better than an earlier design that prompted GPT-3 to generate an abstracted schema for a given problem statement, taking inspiration from previous work on schema-based analogical transfer \cite{gickSchemaInductionAnalogical1983,yuDistributedAnalogicalIdea2014}. We also explored a range of zero-shot to few-shot prompt-based learning paradigms. The structure of the prompt given to GPT-3 in the experiments in this paper is shown in Figure \ref{fig:prompt}. We constructed three example problem-analogy pairs to illustrate the core idea of a cross-domain analogy. Figure \ref{fig:analogy} shows the three examples we constructed. 

% Below is one of the problem input and analogy output example pairs (reformulated from Duncker’s radiation problem) that we used:
% \begin{quote}

%  \textbf{Input}
 
% Stakeholder: a patient who has a malignant tumor in his stomach 

% Context: ray at low intensity is insufficient to destroy the tumor 

% Goal: destroy the tumor without affecting the healthy tissue 

% Obstacle: ray at high intensity will also destroy healthy tissue

% \textbf{Output}

% Stakeholder: a city ruled by a fortress of invaders

% Context: many roads led to the fortress

% Goal: capture the fortress with minimum injuries and deaths 

% Obstacle: invaders have planted mines in the main road
% \end{quote}

% \begin{quote}
%     I want you to generate a new situation that has a very different x (stakeholder, context) but very similar y (goal, obstacle) as the output.

%     [examples of input, output pairs as shown in Figure \ref{fig:analogy}]

%     Input
    
%     [target input]

%     Output

% \end{quote}

We wrote six design problems across various topics (food insecurity, job security, entertainment, etc) as shown in Figure \ref{fig:6-problems}. Figure \ref{fig:6-problems} also shows examples of prompt and GPT-3 generated output pairs with zero-shot, one-shot, and few-shot paradigms.
% The list of design problems are attached in Appendix \ref{appendix:b}. Examples of prompt-output pairs are shown in Figure \ref{fig:gpt3-analogy}.
We used GPT-3 to generate 20 analogies each for input problems, for each of three prompt programming paradigms: zero-shot - no example of input and output; one-shot - one example; and few-shot - three examples. To reduce the potential for overfitting on a specific example in the one-shot case, we conducted two runs of the one-shot paradigm, each with a different example (the \texttt{manager} and \texttt{Duncker} analogy examples). This process resulted in a total of 480 problem-analogy pairs across four runs (one zero-shot, two one-shot, and one few-shot). %In the one-shot paradigm, we used either \todo{the Duncker analogy} or \todo{the manager analogy}. In the few-shot paradigm, we used all three examples in the prompt. 
The code for these prompts, along with the 480 generated analogies, is included in the Supplementary Material.

\subsection{Measures}

All 480 analogies generated with different methods were rated by a single PhD-level researcher with expertise in creativity support tools, example-based creativity, and problem reformulation, who was blind to condition. Analogies were rated for potential usefulness for creative reformulation. The primary criteria were 1) whether there was an analogical mapping between the generated analogy and the design problem, 2) whether the analogy was unique or a duplicate/repeat of any other generated analogy for the same problem, and 3) a ``best guess" of whether the potential inferences from the analogy might be useful for inspiring new creative formulations of the problem. Note that the ``potential usefulness" criterion would be more systematically and robustly investigated in Study 2; here, we use these heuristic judgments to inform our initial prompt engineering efforts, before systematically testing the outputs from the best-performing prompt design in the context of a creative task. The researcher also measured the degree of duplication manually during screening, judging whether analogies were so similar to previously encountered analogies that it was redundant. To summarize, the rater screened the analogies and categorized them using three codes: 1) not potentially useful, 2) potentially useful and new, and 3) potentially useful but duplicated with a previous one. Examples of duplicate analogies are shown in Figure \ref{fig:dup-analogy}.

Because we are interested in the ability of LLMs to help with cross-domain analogical reformulation, we also measured the similarity of the generated analogies to the input problems, using
% cosine similarity calculated with NLTK\footnote{https://www.nltk.org/}, 
semantic similarity calculated with SentenceTransformers\footnote{https://www.sbert.net/}. 

%To give an intuition for how similar these duplicates were, the cosine similarity between duplicates was <how much>

\subsection{Exploration results}

As shown in Table \ref{tab:pre-screening}, our key result was that one-shot outputs were judged to be consistently more potentially useful ($\sim$80\% compared to $\sim$40\%) and dissimilar (<0.5 compared to >0.7) from the input problems compared to zero-shot outputs. Few-shot outputs were also substantially less similar to input problems ($\sim$0.3 compared to $\sim$0.7) compared to zero-shot outputs, but the judged potential usefulness was slightly lower than one-shot outputs ($\sim$67\% compared to $\sim$80\%. 

While we found an advantage of the one-shot paradigm over the zero-shot paradigm in terms of potential usefulness and semantic distance, we are not confident that the one-shot paradigm would consistently produce better results than few-shot paradigms in general, given the limited number of examples tested and the strong prior from the literature that in-context learning performance improves with the number of examples \cite{brownLanguageModelsAre2020,winataLanguageModelsAre2021}. We \textit{are} comfortable concluding for our study that few-shot learning (with k of at least 1) is likely to yield better results than zero-shot learning; practically, because the one-shot learning paradigm required fewer examples, reducing the effort and cost associated with prompt size, we systematically evaluated the outputs of the one-shot paradigm in the context of a creative task in Study 1.

\begin{table*}%[b]
    \begin{tabular}{|l|l|l|l|l|l|}
        \hline & 
        % \textbf{Avg. cos sim} & 
        \textbf{Avg. semantic} & \textbf{Potential} & 
        \textbf{Duplicate} & \textbf{Unique \& potential}  \\
        &\textbf{sim}&\textbf{usefulness rate}&\textbf{rate}&\textbf{usefulness rate}\\\hline
        Zero-shot & 
        % 0.62 &
        0.72&0.68& 0.24& 0.43 \\ \hline
        One-shot (manager) & 
        % 0.37 &
        0.48 & 0.82 &0.12& 0.70 \\ \hline
        One-shot (Duncker) [for verification] & 
        % 0.39 & 
        0.43 & 0.8 &0.03& 0.77  \\ \hline
        Few-shots &
        % 0.37& 
        0.31 & 0.67 &0.02& 0.65 \\ \hline
    \end{tabular}
    \caption{Results of prompt engineering explorations: Average semantic similarities between original problems and GPT-3 generated analogical problems and human judgments of potential usefulness and duplication. In these runs, the one-shot learning paradigm runs produced the best combination (with an extra round with another example for verification).}
    \label{tab:pre-screening}
\end{table*}

\section{Study 1: Analogy Generation Quality in Usage in a Creative Problem Reformulation Task}
\label{sec:study2}
\begin{figure*}
\centering \includegraphics[width=0.8\textwidth]{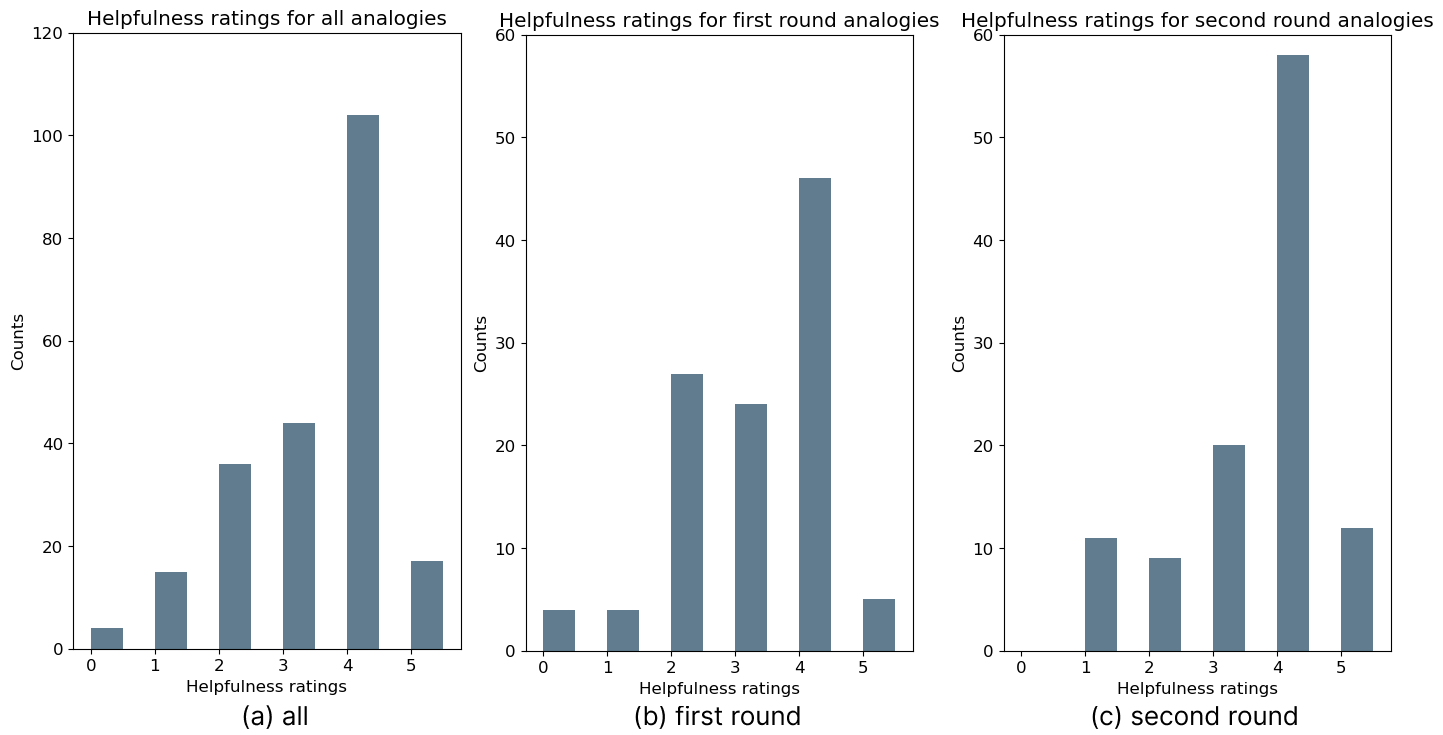}
\caption{Histograms for the helpfulness ratings provided by participants for analogies in the context of the problem reformulation task in Study 1: a) all analogies, b) analogies in the first round, and c) analogies in the second round.}
\label{fig:hist}
\Description{TODO}
\end{figure*}

\begin{figure*}
\centering \includegraphics[width=0.9\textwidth]{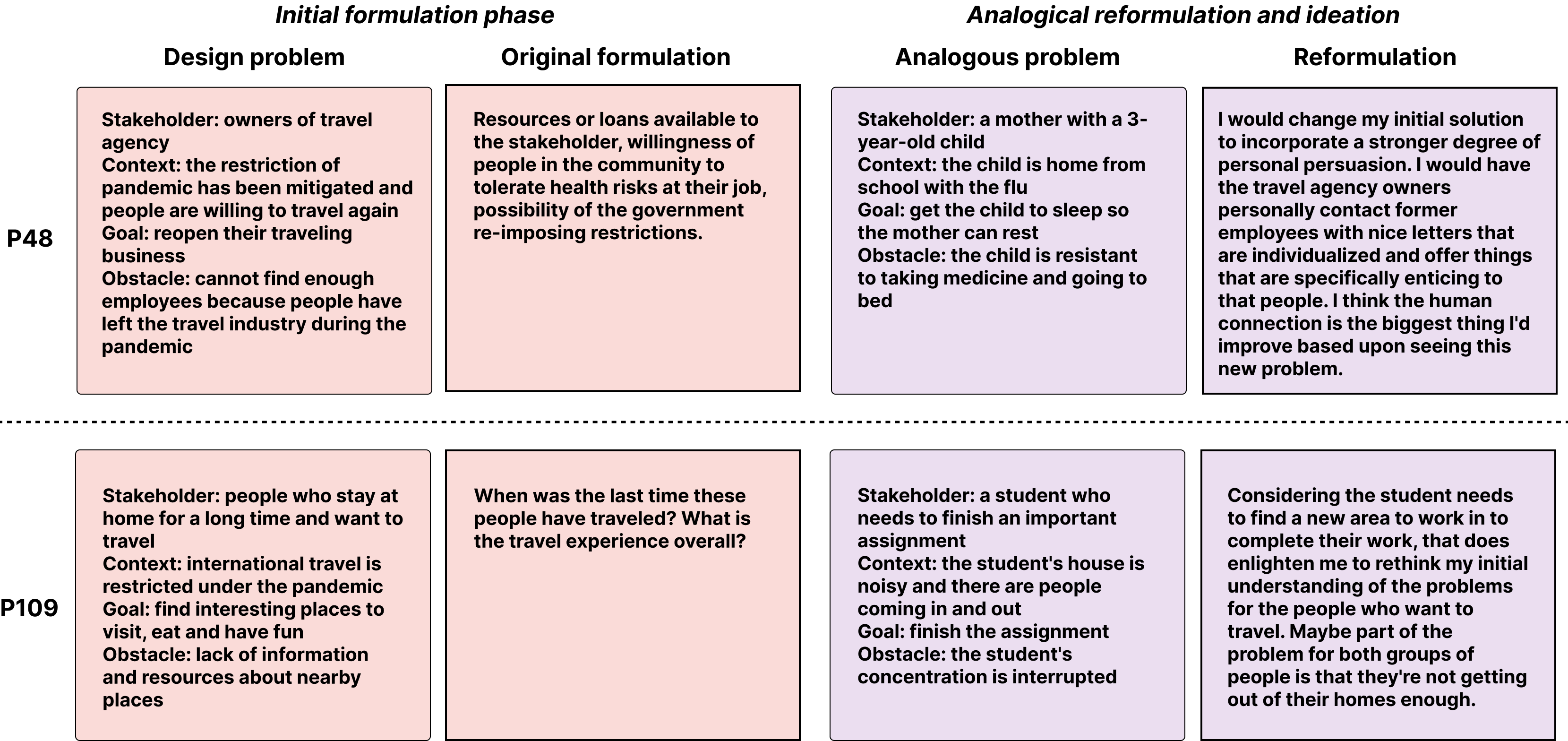}
\caption{Examples of how GPT-3 generated analogies were used for reformulation by two participants (P48, top; and P109, bottom). We show the progression from their original formulation of the problem (from the initial formulation phase) to the reformulation of the problem in response to an analogy (from one of the analogical reformulation phases).}
\label{fig:usage}
\Description{TODO}
\end{figure*}

Having established an initial estimate of the rate at which GPT-3 is able to produce potentially useful cross-domain analogies (i.e., ~70\% unique and potentially useful cross-domain analogies in the best prompt design), we now directly study the degree to which these analogies could actually be useful in a creative task. To do this, we provided the analogy outputs from the best-performing one-shot learning paradigm (i.e., the prompt that included the Duncker example analogy) to people doing a creative ideation task, and asked them to try to use it to inform/inspire their creative ideation and evaluate the analogies' usefulness for that purpose.

\subsection{Methods}

\subsubsection{Design and Materials}
To account for potential variation across people in their reactions to the analogies, we sought to obtain two judgments of usefulness per analogy. We also aimed to minimize the anchoring of our results to any particular design problem (since there was some variation in the duplicate and potential judged usefulness rate across the design problems). Finally, we wanted to balance the time requirements per rater against minimizing the potential overhead of learning the task, since using analogies for problem reformulation may be different from a more typical NLP evaluation task like judging whether a given text is human-like (though still well within the range of everyday ability). Given these requirements, we designed our overall rating task to have two analogies for each rater for the same design problem, and approximately equal representation of analogies across the six design problems. The details of the rating task are explained in the next section. 

To construct sets of two analogies for each rater, we first took the 92 of the 120 total analogies from the Duncker run of the one-shot learning paradigm in prompt engineering explorations that were judged to be both nonredundant and potentially useful, as well as the 17 analogy outputs that were judged to be nonredundant cross-domain analogies but unlikely to be useful. We ignored the remaining 11 analogy outputs that were potentially useful but duplicates (N=4), highly similar to the original problem (N=4), and off-task (e.g., repeating the original problem, incorrect format; N=3). This resulted in 109 unique analogies across the six problems. To enable construction of pairs of analogies, we then randomly selected one of the analogies from a design problem that had an uneven number of usable analogies to drop from the sample. This resulted in a final set of 108 analogies. From these 108 analogies, we then constructed all possible pairwise combinations of analogies within each design problem, and sampled iteratively from these combinations to select pairs, such that each of the 108 analogies showed up in exactly 2 pairs. This resulted in 108 analogy pairs that were given to raters.

% We sampled 108 (90 pre-screened as usable, 18 pre-screened as unusable) unique analogies generated from one-shot learning for user evaluation. We randomly grouped those problem-analogy pairs into 108 pairs of one original problem + its two analogies. Each analogy appeared twice in those 108 pairs.

\subsubsection{Task and Procedure}

% The procedure of Study 1 was illustrated in Figure \ref{fig:procedure}. 
The overall rating task was designed to follow substeps of analogical problem reformulation, beginning with initial formulation of an incompletely specified problem, initial ideation, processing of an analogical stimulus in relation to the problem, and then reformulation and ideation over the reformulated problem. As an example, a designer given a problem of designing ways for people who want to find interesting places to visit and have fun but lack information and resources about nearby places might (re)formulate the problem by \textit{adding} information about the stakeholder(s) (e.g., thinking about friends and family who might have good recommendations) or \textit{rejecting/shifting} a potentially implicit assumption that they need to discover new places, and instead explore new goals around re-experiencing familiar places. This reformulation might be spurred by comparison to analogous problems, such as a parent trying to find ways to entertain their toddler during summer vacations (e.g., by considering how children can often repeatedly enjoy similar experiences or toys in a variety of ways).

Thus, the overall task was divided into three phases: an \textbf{initial formulation} phase, and two \textbf{analogical reformulation and ideation} phases, one for each analogy in the pair assigned to the participant.
% (see Figure \ref{fig:procedure} for an overview).
In the \textbf{initial formulation} phase, participants were first given a design problem and asked to construct an initial formulation of the problem through adding details to the problem components (stakeholder/context/goal/obstacle), in response to the question, ``What characteristics of the stakeholder/context or other problem components do you think would be important to consider when trying to solve the problem?". Then, participants were asked to generate at least one solution to that problem with details added. In each of the two \textbf{analogical reformulation and ideation} phases, participants were presented with an analogy, and then answered two questions 1) ``What interesting commonalities and differences do you notice between the problems (the original and analogical problems)?" 2) ``Based on your comparison of the original problem with the new problem, can you think of ways to change your initial understanding of the original problem that might lead to new solutions? What might you add/change to your understanding of the problem components?" The first question was meant to facilitate analogical/case comparison, a known effective strategy for inducing analogical schemas for ideation \cite{gickSchemaInductionAnalogical1983,loewensteinHowOneHook2010,kurtzComparisonPromotesLearning2013}. The second question was meant to probe for potential \textit{reformulation} of the problem based on the analogical mapping. After this, participants were asked to come up with a new solution to the original problem. To get a more realistic sense of the potential value of the analogies for ideation, we did not force participants to generate ideas in this phase from the analogies: participants were allowed to use Google to search for inspiration to generate ideas. After each task, participants were asked to rate how helpful they found it to compare their problem situation with the analogous one from 1 -  ``Not helpful at all" to 5 - ``Very helpful" (no rating was 0). This process was then repeated for the second analogy in the set. After the second analogical reformulation and ideation phase, participants had a chance to give open-ended comments on the study. Appendix \ref{appendix:a} shows screenshots of the task interface for the initial formulation and analogical reformulation and ideation phases of Study 1. We obtained institutional IRB approval for the  project prior to Study 1.
% participants were asked to do additional rounds of reformulation and ideation with problem analogies pre-generated by language models. In each round the participants were asked 

\subsubsection{Participants} 
\label{sec:participant}

We recruited participants from Prolific\footnote{https://www.prolific.co/} as Prolific participants typically exhibit greater care in completing tasks compared to Amazon MTurk participants \cite{oppenlaender_creativity_2020}. We limited all participants to U.S. residents. Each participant was paid \$5.25 (\$10.5 per hour) for their participation of 30 mins. %It makes sense to recruit crowd workers for this purpose because most people have the basic linguistic competence of noticing when a problem has been reformulated. 
A total of 123 participants enrolled the task, but 13 of them only finished the first round of analogical reformulation and ideation and did not complete the second round. After removing those participants, our final sample included 110 participants who finished the study. Due to our randomized procedure for assigning participants to analogy sets, removing incomplete responses also resulted in 22 analogies being used and evaluated by only one participant.

% \begin{figure}
% \centering \includegraphics[width=0.65\textwidth]{figures/procedure.png}
% \caption{Procedures of Study 1: one phase of initial problem formulation and ideation, and two rounds of problem reformulation and ideation attempts with GPT-3 generated analogies for the design problem.}
% \label{fig:procedure}
% \Description{TODO}
% \end{figure}
\subsubsection{Systematic analysis of reformulation behaviors}
\label{sec:coding}

To understand how participants used the analogies to reformulate the original problems, we systematically coded participants' responses to the reformulation question in the analogical reformulation and ideation phases:

\begin{quote}
``Based on your comparison of the original problem with the new problem, can you think of ways to change your initial understanding of the original problem that might lead to new solutions? What might you add/change to your understanding of the problem components?''
\end{quote}

We wanted our coding scheme to capture a core distinction in the creativity literature between formulation and reformulation. The former --- formulation --- entails adding details to to transform an ill-structured problem into a more structured problem that can make the search for solutions more tractable. This process is necessary because creative problems are not static “givens”; rather, these problems are ill-structured and malleable. In contrast to well-structured problems (e.g., the Tower of Hanoi, algebra), where the initial description of the problem sufficiently delineates the goal state and allowable transformations, in creative problem domains, such as innovative design or interdisciplinary research, the “problem statement” typically underspecifies what the desired solution is. In information-processing terms, “the start state is incompletely specified, the goal state is specified to an even lesser extent, and the transformation function from the start to goal states is completely unspecified” \cite{goelStructureDesignProblem1992}. Analogical transfer can be instrumental in this elaboration process, adding functional decompositions or subproblems to solve that might address the high-level problem \cite{helmsCompoundAnalogicalDesign2008}. The latter --- reformulation --- entails reshaping an existing structured problem, often by rejecting or relaxing constraints and assumptions, or even reformulating the goal itself. An innovative designer can choose to see the design problem differently, to relax certain constraints or redefine certain customer needs in other terms, thereby arriving at innovative solutions. Such problem-reframing has been characterized as key element of creative expertise in design \cite{crossExpertiseEngineeringDesign1998, schonReflectivePractitionerHow1983}. For example, Bucciarelli \cite{bucciarelliDesigningEngineers1994} observed in his ethnographic studies of engineering designers that a common strategy was to re-evaluate and sometimes alter formulations of the design problem. Goel and Pirolli \cite{goelStructureDesignProblem1992} call this a “reversal of the transformation function”, where, instead of reducing the difference between the current and goal states by means of a design “move” (e.g., a particular product configuration), a designer moves closer to the goal state by changing what the goal state is. In insight problem solving, too, a key mechanism for moving past design impasses, is to relax constraints and assumptions \cite{knoblichConstraintRelaxationChunk1999}, or even change what details of the problem are attended to, which can be characterized as a search not for solutions, but for problem spaces, where solutions are more readily found \cite{kaplanSearchInsight1990}. Here, too, analogy can be used to support reformulation; For example, \cite{dorstFrameInnovationCreate2015} described a case where Designers used an analogy to a music festival to (re)frame the problem of night violence in King's Cross from a crime prevention/reduction problem to an entertainment frame.

Thus, we initially employed two codes - "adding" and "shifting" - to capture notions of "formulation" (elaborating/adding), and "reformulation" (changing/rejecting/shifting). However, we discovered that the "shifting" code also involved instances of "adding," and disagreements arose regarding whether a reformulation constituted "adding," "shifting," or "adding" plus "shifting." This resulted in low inter-rater reliability. To address this, we decomposed the "shifting" code into two more concrete and reliable codes: "adding" and "rejecting", which were not mutually exclusive, and (if both are present) could still capture the notion of shifting. Our final coding scheme thus consisted of two independent codes: 1) \textbf{adding} (0/1), indicating any presence of \textit{new} concepts or descriptions in the answer compared to the initial problem statement or formulation; and 2) \textbf{rejecting} (0/1), indicating direct statements from the participant that they are rejecting a goal/obstacle/context/ stakeholder from the problem or their initial formulation. A problem reformulation could be seen as a combination of adding and rejecting. Examples of adding and rejecting are provided below:

\begin{quote}
\textbf{Example of adding}

\textit{Original problem}

Stakeholder: people who stay at home for a long time and want to travel

Context: international travel is restricted under the pandemic

Goal: find interesting places to visit, eat and have fun

Obstacle: lack of information and resources about nearby places

\textit{Analogy}

Stakeholder: a software company

Context: developing a new mobile application

Goal: find out user requirements

Obstacle: most of the users are not familiar with the application's purpose

\textit{Participant’s response to reformulation question}

I would \textbf{add specific locations} to the problem because that would narrow down the scope of \textbf{how to get information to them}.
\end{quote}

\begin{quote}
\textbf{Example of rejecting}

\textit{Original problem}

Stakeholder: owners of travel agency

Context: the restriction of pandemic has been mitigated and people are willing to travel again

Goal: reopen their traveling business

Obstacle: cannot find enough employees because people have left the travel industry during the pandemic

\textit{Analogy}

Stakeholder: a farmer

Context: the restriction of the use of pesticides has been mitigated

Goal: use pesticides to increase crop yield

Obstacle: the farmer cannot afford to buy pesticides

\textit{Participant’s response to reformulation question}

\textbf{It's not that the travel agency can't find employees, it's that they can't afford to pay employees to work for them after being closed for so long}, thus causing a feedback loop of: not enough employees -> less money ->cant afford to hire employees -> not enough employees.
\end{quote}

Two graduate-level researchers with expertise in creativity support tools, example-based creativity, and problem reformulation, applied the coding scheme to the participants' reformulation responses.
After five rounds of coding, discussing (and resolving) disagreements, and refining the coding scheme between two expert coders, the inter-rater reliability reached Cohen’s Kappa=1 (perfect agreement) for adding codes and Cohen’s Kappa=0.847 (near perfect agreement) for rejecting codes respectively for 50 problem-analogy pairs \cite{landis1977measurement}. After developing the substantial inter-rater reliability, one coder coded the remaining 170 reformulation responses.

\begin{table*}[]
\begin{tabular}{|l|l|l|l|l|l|l|l|}
\hline
 % &&	&	&	&	&	& Median\\
&Count&Median of helpfulness rating\\ \hline
Adding	&185 (84.09\%) &4\\ \hline
Rejecting	&23 (10.45\%)&4	 \\ \hline
Adding + Rejecting	&22 (10\%)&4	 \\ \hline
Total	&220&4	 \\ \hline
\end{tabular}
\caption{Systematic coding results (adding / rejecting / adding + rejecting) of analogical reformulations in Study 1.}
\label{tab:coding}
\end{table*}

\subsection{Quantitative Results}

All 220 analogy-response pairs from all completed participants (N=110) are included in the Supplementary Material. We first report a range of quantitative analyses of LLM-generated analogies' quality at a whole sample level, before describing some illustrative case studies of how the analogies were processed and used to contextualize the high level results.

\subsubsection{Median 4 out of 5 helpfulness rating for analogy, with increased rating on the second analogy}

Since analogies received unequal numbers of ratings, we report the median as an estimate of the central tendency of ratings across analogies. The median rated helpfulness of analogies was 4 out of 5 (see Table \ref{tab:coding} and Figure \ref{fig:hist}). Interestingly, we noticed that ratings were higher for the analogies that were rated second in the analogical reformulation phase ($median = 4$) compared to the analogies that were rated first ($median=3$). A Wilcoxon test (within subjects, two groups) estimated that this increase in median ratings from the first to second analogy was statistically significant ($statistic=561.5, p-value=0.002$). We also compared the median of helpfulness ratings for analogies judged by two or more participants and the median of helpfulness ratings for all analogies to check whether there was a difference between only one judgment vs. multiple judgments: both medians were 4, suggesting that we obtained similarly reliable estimates from single vs. multiple-judgments for each analogy

%\subsubsection{Correlation between helpfulness ratings, a prior ratings, and add/reject/any change codings}

\subsubsection{Frequent analogical reformulations, primarily by adding information, positively correlated with helpfulness ratings}

The results of our systematic coding of the reformulation responses are reported in Table \ref{tab:coding}, showing 185/220 (84.09\%) instances of adding, and 23/220 (10.45\%) instances of rejecting. There was high overlap between instances of adding and rejecting: specifically, there were 22/220 (10\%) instances of both adding and rejecting (shifting).

Since the add/reject/any change codings were binary and the helpfulness ratings were continuous, we calculated point-biserial correlations between them. We observed a statistically significant positive correlation between reformulation responses coded as having an addition and helpfulness ratings for the corresponding problem-analogy pair ($point-biserial = 0.2559, p = 0.0001$) and between any change codings (add or reject) and helpfulness ratings ($point-biserial = 0.2669, p = 0.00006$). However, we observed no statistically significant correlation between reject codings and helpfulness ratings ($point-biserial = 0.0094, p = 0.8891$).

% Given that the instances where reformulation was not happening were not meaningful chunks of text to compare against the formulation, we restricted our analysis to cases where we agreed that some change was happening. Our goal was to describe the nature of the change in terms of similarity to the original formulation. We compared the distributions of similarity for adding vs. rejecting forms of reformulation as shown in Figure \ref{fig:transformer_sim}.

% \begin{figure}
% \centering \includegraphics[width=0.5\textwidth]{figures/Cos_sim.png}
% \caption{Probability density function of cosine similarity between formulation and reformulation for adding (green) and rejecting (red).}
% \label{fig:transformer_sim}
% \Description{TODO}
% \end{figure}

\subsubsection{No correlation between a priori usable ratings and in-use helpfulness ratings}

To explore the correspondence between the \textit{a priori} judgments of potential usefulness from prompt engineering explorations (outside the context of a creative task) with in-use ratings of usefulness of analogies, we calculated the correlation between the potential usefulness ratings from prompt engineering explorations and the helpfulness ratings here. Since the potential usefulness ratings were binary and the helpfulness ratings were continuous, we calculated a point-biserial correlations between them. We observed no statistically significant correlation between \textit{a priori} usable ratings and helpfulness ratings: $point-biserial = -0.0266, p = 0.6939$. However, since we only utilized one researcher's ratings to predict the usefulness ratings provided by another set of participants, we cannot discount the possibility that the lack of correlation might be due to alternative explanations. For instance, it could be that the specific rater was unable to predict usefulness without actually using the analogies, or that the difficulty of predicting usefulness for someone else contributed to the lack of correlation.

\subsection{Qualitative descriptions of how people used LLM-generated analogies to reformulate problems}

To complement our systematic quantitative analyses of participants' reformulation behaviors in \ref{sec:coding}, we illustrate how participants used analogies with two more cases shown in Figure \ref{fig:usage}. In the first example of the ``local tourism" problem, P48 wanted to gain more information of the context such as external resources and constrains ("resources or loans", ``willingness of people", etc). P48 described additional contexts that were not included in the original design problem, such as ``resources or loans that were available'' and ``willingness of people... to tolerate health risks.'' After seeing the analogical problem that a ``child is resistant to taking medicine and going to bed", P48 highlighted the role of persuasion to solve the original problem and incorporated personal persuasion into the solution. P48 did not explicitly reformulate the original problem, but instead began ideating a solution, which is a common behavior for novice \cite{macneilDesigningCrowdCognitive2019, macneilFindingPlaceDesign2021}. But the solution suggested that P48 noticed a potential core obstacle ``a lack of personal persuasion" within the original problem that employers ``cannot find enough employees because people have left the travel industry during the pandemic". This is interesting because ``the child is resistant to taking medicine and going to bed" can be seen as an interesting and plausible analogy to ``previous employees of the travel industry who are resistant to return". In the second example, the original problem formulation of P109 was around the stakeholder. After seeing the analogy, P109 noticed the high-level similarity between ``lack of entertainment information for people who stay at home" and ``noisy environment of a student's house" and then explicitly reformulate the root problem as ``they're not getting out of their homes enough". Those two examples demonstrated that the generated analogies were able to facilitate participants to deepen their understanding of the original problem and reformulate the root issues, which could inform better solutions.  

Complementing these example cases are some notable open-ended comments from participants about how the analogies impacted their formulation and ideation processes.
% \subsubsection{Participant comments on analogical examples and thinking process}
For example, P40, P29, and P109 commented on how the analogies opened their mind:

\begin{quote}
Comparing the two problems \textbf{opened new mental doors}, only one of which led to a practical, workable solution. But the fact that those mental doors were completely invisible to me before seeing the analogous problem suggests that looking at related but not identical problems can be extremely helpful. (P40)
\end{quote}

\begin{quote}
I think reading from analogies can definitely \textbf{open your mind up} to new ideas and improve critical thinking skills. I didn't even realize at first that it was ``people" in Philadelphia, until the next slide. I totally overlooked it, but it's interesting how so many things correlate, even if they're very different. (P29)
\end{quote}

\begin{quote}
This was extremely interesting for me! I really like that I was able to compare problem situations with the analogies here and expand my original thinking. This allowed me to identify new, creative solutions to the original problem that I would not have thought of without the expanded analogies!
(P109)
\end{quote}

% There were 26 participants explicitly giving positive feedback on this task in the optional comment space, comparing to only 6 participants explicitly giving negative feedback. Some participants said that analogies were helpful: “Having other examples is really helpful! Thank you so much, I thoroughly enjoyed this survey!” (P31). 

% Other participants indicated that they enjoyed the thinking process and exercise in general: “This was a really interesting way to think about things, I enjoyed it.” (P14); “Thank you! This was an interesting thought exercise and a great approach for generating ideas” (P71). P89 said “I don't think I have ever used analogous problem solving in this way. This was a very interesting exercise.” P15 even said that he or she wanted to learn that kind of analogical thinking:

% \begin{quote}
% this was actually fun to me. I didn't google anything because I didn't want to be swayed by anything I cldv read. Thank you so much for the opportunity!! Now I'm interested in learning more about this way of thinking.
% \end{quote}

% More specifically, P109 mentioned how the analogical exmples benefited creative thinking process:

Similarly, P79 noted: %Even the main method for analogical thinking - abstracting core concepts by comparing the original statement and analogy - was mentioned by a participant:

\begin{quote}
These \textbf{juxtapositions} create very higher order thinking involving public policy expertise as well as scientific and community knowledge. Very interesting! These creative solutions are necessary to improve social welfare and environmental problem solving. Thank you! (P79)
\end{quote}

Another participant's comment also illuminated the lack of correlation between pre-judged potential usefulness and helpfulness in the context of the creative task:

%On the other hand, a few participants said it was hard to think with analogies “This survey was kinda difficult or i'm just not that bright” (P102); or the analogies were random 

\begin{quote}
    Overall, I found the analogies to be a bit random and not the best fit, but they did make me think of the problem from a slightly different angle, which was interesting.” (P68).
\end{quote}

Overall, these qualitative descriptions and open-ended comments complement the quantitative results above by adding more concrete details to illustrate \textit{how} the LLM-generated analogies were useful in the creative problem formulation task.

\section{Study 2: Potential Harmfulness and Toxicity of Generated Analogies}
\label{sec:study3}
\begin{figure*}
\centering \includegraphics[width=0.9\textwidth]{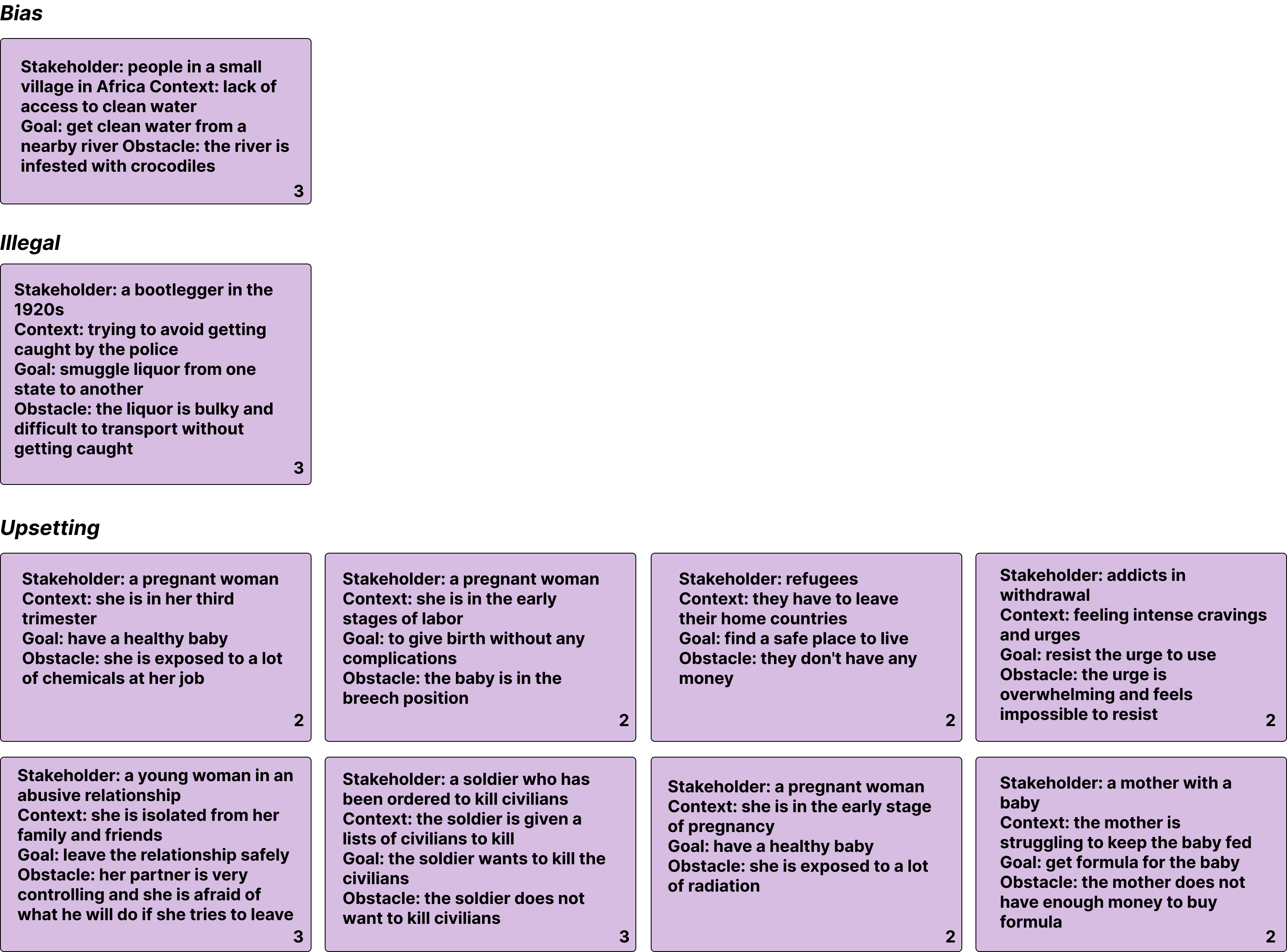}
\caption{Analogies that were flagged by at least 2 of the 5 raters as potentially harmful under three categories: biased, illegal, and upsetting. The subscript for each example showed how many people flagged the example as harmful.}
\label{fig:harmful}
\Description{TODO}
\end{figure*}

In this final section, we systematically investigate the degree to which GPT-generated analogies might be harmful or toxic. We do this because we are interested in real-world usage of a method like this, and want to facilitate weighing of cost-benefit considerations and design of mitigations for potential harms, especially given concerns about potential harms from large language model usage \cite{benderDangersStochasticParrots2021}.

Our analysis here is anchored on the potential use case of giving LLM-generated analogies to designers as an aid to problem formulation. Inspired by previous analyses of the ethics of LLMs based not just on direct harm to the user but also participation in harmful and oppressive systems of power \cite{blodgettLanguageTechnologyPower2020,benderDangersStochasticParrots2021}, we consider the following primary categories of harm: 1) perpetuating hegemonic systems of oppression via uncritical incorporation of harmful stereotypes and biases in LLM outputs, 2) unknowingly framing problems in a way that leads to solutions that violate laws, and 3) experiencing psychological distress from upsetting or abusive content.

Analyzing the bias and stereotype-based categories of harm typically involves consideration of specific social categories at risk of harm from biases, such as gender \cite{shengWomanWorkedBabysitter2019}, or disability \cite{hutchinsonSocialBiasesNLP2020}. As Bender et al \cite{benderDangersStochasticParrots2021} note, which particular social categories are salient is culture-bound. Thus, we note here that we anchored our conceptualization of protected social categories from a US-centric perspective due to our positionality as US-based researchers, and considered the dimensions of disability, race/ethnicity, gender, sexual orientation and age. To analyze potential illegal suggestions, we considered the context of US laws, again due to our positionality. Finally, we intentionally drew on raters from a variety of demographic backgrounds of race/ethnicity and gender, to allow for more expansive detection of content that may be considered psychologically distressing. We note all these to emphasize the situatedness of our results, and make no claims as to their generality to other contexts, and share our raw outputs to assist with extensions of this work that consider this question from other positionalities.

From a measurement perspective, harmfulness audits have often been done using automated measures such as the Perspective API\footnote{https://www.perspectiveapi.com/} toxicity classifier \cite{benderDangersStochasticParrots2021}. However, the sensitivity and accuracy of these measures is uncertain, and there is evidence that they may produce false positives with mere mentions of terms associated with marginalized identities, such as race or disability \cite{jigsawUnintendedBiasIdentity2018,hutchinsonSocialBiasesNLP2020}. Thus, we chose to manually review all outputs, and estimate ranges in rates of harmfulness or toxicity, as well as qualitatively describe patterns of outputs on these dimensions.

\subsection{Methods}

Three members of the research team (1 White male junior faculty, 1 Asian immigrant male junior faculty, and 1 Asian immigrant male PhD student), and two additional research assistants (1 Asian female undergraduate student, 1 North African female PhD student) reviewed all 108 analogies used in Study 1. We report the general demographics of the screeners to facilitate understanding of how our results here on the potential harmfulness of the outputs may be shaped by our positionality, recognizing that other researchers or users in different positionalities may reach different conclusions about the LLM outputs.

% Drawing on... We conceptualized harmfulness broadly as 1) encoding biases/stereotypes against protected social categories, such as race or gender, 2) including abusive language, 3) depicting illegal behavior, or 4) including upsetting content. We anchored our conceptualization of protected social categories from a US-centric perspective due to our positionality as US-based researchers, and considered the dimensions of disability, race/ethnicity, gender, sexual orientation and age. 

Using the three-fold conceptualization of harm described above (bias, illegality, and potential psychological distress), each screener independently went through each analogy, and flagged it as potentially harmful or not. Screeners also added descriptive comments about potential harm, where appropriate. We use the \textit{harmful}/ \textit{not harmful} flags to estimate rates of harmfulness, and the descriptive comments to qualitatively describe the nature of potential harms.

\subsection{Results}

There was wide variation in the number of flagged analogies across screeners, ranging from 1 to 27 out of 108 total analogies: specifically, the number of flagged analogies across the five screeners was, in ascending order: 1, 2, 2, 10, and 27. Using the most expansive approach of counting any flag by any screener increased the upper bound rate to 28/108 (26\%). 

Qualitatively, the majority of the analogies that were flagged across screeners were for potentially upsetting content, rather than biased/abusive language or illegal behavior. Figure \ref{fig:harmful} shows the 10 analogies that were flagged by at least 2 of the 5 raters as potentially harmful, along with the associated screener notes. There was one clear instance of biased depiction of people in a small village in Africa with multiple stereotypes of Africa as poverty-ridden and a "wild safari". There was also one instance of illegal behavior described (smuggling liquor in the 1920's). The remaining 8 analogies were flagged for potentially upsetting content, such as difficult birthing/pregnancy, refugees fleeing their country, trying to escape an abusive relationship, substance addiction, and war. In the subset of screening results from the screener who flagged the most analogies, we also observed a similar pattern where predominantly potentially upsetting content accounted for most of the potentially harmful analogies.

% LLM-generated analogies, as a part of design material, should avoid or reduce harmful content. As a first step, we measured the harmfulness of those generated analogies. Three HCI experts (two professors, one PhD students) with various ethnicity and cultural backgrounds evaluated the harmfulness of 108 generated analogies with majority voting, focusing on toxic/stereotypical/illegal content. The majority voting process resulted only three analogies, as listed below:

% \begin{quote}

% \textbf{Harmful analogy 1 - harmful stereotype (african villages = rural, lack clean water)}

% Stakeholder: people in a small village in Africa

% Context: lack of access to clean water

% Goal: get clean water from a nearby river

% Obstacle: the river is infested with crocodiles

% \end{quote}

% \begin{quote}

% \textbf{Harmful analogy 2 - violence}

% Stakeholder: a soldier who has been ordered to kill civilians

% Context: the soldier is given a lists of civilians to kill

% Goal: the soldier wants to kill the civilians

% Obstacle: the soldier does not want to kill civilians
% \end{quote}

% \begin{quote}

% \textbf{Harmful analogy 3 - illegal activity}

% Stakeholder: a bootlegger in the 1920s

% Context: trying to avoid getting caught by the police

% Goal: smuggle liquor from one state to another

% Obstacle: the liquor is bulky and difficult to transport without getting caught

% \end{quote}

\section{Discussion}
\subsection{Summary and interpretation of findings}
\label{sec:dis_result}
In this paper, we directly investigated the potential of LLMs to generate analogies that might be useful for creativity support, specifically in the task context of augmenting cross-domain analogical reformulation. Through carefully crafted prompts, we were able to use prompt-based learning to generate cross-domain analogies with GPT-3 that were semantically distant from source problems and judged to be potentially useful at a high rate ($\sim$70-80\% of outputs; prompt engineering explorations). We complemented this prompt engineering exploration with a systematic analysis of how people used LLM-generated analogies in the context of a creative problem reformulation task, and found that they were able to leverage the analogies to reformulate design problems ($\sim$80\% of cases), rated the analogies overall as helpful for reformulating the problems (average $\sim$3-4 Likert rating out of 5), and described the analogies as augmenting/broadening their thinking process (Study 1). Investigating potential tradeoffs against toxic/bias/harmful outputs, we found an upper bound of $\sim$25\% of outputs being potentially harmful in some way, with a clear majority ($\sim$80\%) of such cases being judged as such due to descriptions of potentially upsetting situations, rather than biased or toxic descriptions of specific social categories. 

We infer from these analyses that LLMs, with carefully crafted prompts, can frequently generate analogies that can augment cross-domain analogical reformulation. The cost/benefit analysis of deploying these analogies in creativity support tools may vary by situation given the potential for upsetting content. However, note that our screeners were emphasizing minimizing false negatives, and many problematic situations may inherently contain content that may be upsetting to some (since they, by definition, describe situations that people want to transform to ``better" states \cite{newellHumanProblemSolving1972,rittelDilemmasGeneralTheory1973,dorstFrameInnovationCreate2015}).

More broadly, our results add to prior work on the analogical reasoning capacities of LLMs, extending from simpler analogy word problems \cite{brownLanguageModelsAre2020,webbEmergentAnalogicalReasoning2022,bhavyaAnalogyGenerationPrompting2022} to more complex cross-domain analogical outputs, and extending prior proofs-of-concept \cite{webbEmergentAnalogicalReasoning2022,bhavyaAnalogyGenerationPrompting2022,zhuGenerativePreTrainedTransformer2022} with systematic testing of analogical outputs in the context of creative tasks. Our prompting methodology with structured inputs also has the potential to be extended to other domains with repeating structural components, such as scientific reasoning (e.g., theory, evidence, method) or legal reasoning (e.g., precedent, defendant, issue). Additionally, our observed lack of correlation between a priori judgments of potential usefulness from prompt engineering explorations and helpfulness ratings and reformulation behaviors in the context of an actual task underscore the importance of going beyond simple benchmark-based measures of output ``quality" \cite{leeEvaluatingHumanLanguageModel2022}, or even human judgments of quality separate from usage in a creative task \cite{clarkAllThatHuman2021} (although, as we note in the results section for Section \ref{sec:study2}, we cannot rule out alternative explanations such as lack of ability for this specific rater, or the possibility of successfully predicting usefulness for oneself, but not others). The significant increase in helpfulness ratings from the first to second analogy in the analogical reformulation phases in Study 1 also suggest that some degree of ``settling into" a creative analogy-making task is needed to get more accurate measures of the quality of LLM-generated analogies; this effect is reminiscent of prior results showing the induction of a ``relational mindset" from initial processing of cross-domain analogies that improved subsequent recognition and processing of cross-domain analogies \cite{vendettiFarOutThinkingGenerating2014,goldwaterCanRelationalMindset2019}.

% \todo{harmfulness}
\subsection{Limitations and next steps}
\subsubsection{Beyond the one-shot prompt-based learning paradigm}
In this paper, we only sampled from one run of the one-shot paradigm with one example for Study 1's in-use evaluation due to the cost of human evaluation. This made sense given that prompt engineering explorations showed an advantage of the one-shot paradigm over the zero-shot paradigm in terms of uniqueness and judgments of potential usefulness, and comparable uniqueness and potential usefulness to the few-shot paradigm. However, while we are fairly confident that the one-shot paradigm produces better results than the zero-shot paradigm, we are not confident that one-shot paradigms would be consistently better than few-shot paradigms, given the small number of examples we tested, and the substantial literature on how in-context learning improves with the number of examples \cite{brownLanguageModelsAre2020,winataLanguageModelsAre2021}. A potential direction for future work is to expand on different instantiations of few-shot learning paradigms. For example, future work might explore the effect of providing uniform or diverse examples on the usefulness of generated analogies. It may also be fruitful to more systematically explore variations of prompt programming, such as more specific, varied, or longer / more complex prompts, as well as investigating the extent to which a more traditional fine-tuning approach may yield substantially better results. That said, from a practical standpoint, one-shot learning paradigms have the benefit of requiring less examples, which reduces prompt engineering effort and costs associated with prompt size. %One few-shot paradigm does open more prompt programming opportunities: for example, whether a diverse set of input-output analogy examples will improve the usefulness of output analogies or not. Future work is encouraged to explore the effect of longer and more complex prompts such as various combinations and sequences of examples for few-shot learning. Comprehensive prompts have the potential to better connect to the domain knowledge that is encoded within LLMs through fine-tuning with more examples or other modular architectures to strengthen the creativity and expertises of LLMs, which is a promising direction of future work.

\subsubsection{Improving the depth of insight in LLM-generated analogies}
While the rate of generating cross-domain analogies was relatively high, it is unclear to what extent these analogies could spark deep insights or highly creative conceptual leaps. Note that 
% Our studies have shown evidence that Large Language Models (LLMs) are able to understand and conduct the analogy generation task. Particularly, a wide range of concepts, commonsense knowledge and context were observed in the generated analogies. This is not surprising, given that LLMs have processed a huge amount of text information and constructed extensive knowledge. That knowledge embedded in LLM contributed to humans’ understanding of design problems by connecting the current problems with new concepts or considerations, which was reflected in the 84.09\% rate of adding code in problem reformulations. On the other hand, it was also observed 
the rate of \textit{rejecting}/shifting elements of the problem in participants' reformulation (10.45\%) was fairly low. This relative lack of rejecting/shifting may reflect the particulars of the task and participants in Study 1. But we also wonder if, as others have suggested \cite{webbEmergentAnalogicalReasoning2022}, the insightfulness of LLM-generated analogies may be bounded by the degree to which the model has access to rich domain knowledge, such as commonsense or physical knowledge. In this sense, LLMs' capacity for analogy generation observed here may be more a function of what Mahowald et al \cite{mahowaldDissociatingLanguageThought2023} call ``formal linguistic competence" (knowledge of rules and patterns of a given language), as opposed to deeper ``functional linguistic competence" (being able to understand and use language in real-world tasks, such as formal reasoning, situation modeling, and social reasoning. This general idea is consistent with the informal observations of the screener for prompt engineering explorations, who noticed that many of the analogies seemed to be quite diverse, but didn't describe deep details about any particular domain. We wonder what the LLM-generated analogies might inspire if they were generated from a combination of LLMs and more structured/specialized knowledge bases (e.g., modeling common sense, domain ontologies, or physical world modeling); a recent proof-of-concept of something like this is a prototype using ChatGPT as a natural language interface, drawing from the much more structured and curated computational knowledgebase of Wolfram Alpha \cite{wolframWolframAlphaWay2023}. % which could imply that LLMs did not have a deep understanding of the original problems which led to a lack of highly inspiring concepts or ideas in the LLM-generated analogies. Ideally, analogies should focus on deeper domain and world knowledge instead of superficial similarities to invoke “ah-hah” moments, which usually occurs in successful human-generated analogies such as the analogy used for the Odon Device (extract cork from wine bottle -> pull infant through birth canal).

\subsubsection{From static to iterative LLM-assisted cross-domain analogical reasoning}
Our experiments were conducted with a static set of pre-generated examples for the zero- and few-shot prompts. %we simplified our prompts for analogy generation with only instructions and examples to provide essential guidance to LLMs, and we pre-generated all the analogies. 
We wonder how LLM-powered analogy generation might be integrated into the more iterative nature of design and creative cognition  \cite{dorstCreativityDesignProcess2001}. For instance, rather than using a fixed design problem as a source for LLM-generated analogies, could we integrate users' reformulations into subsequent prompts? %we For example, we asked participants’ to formulate and solve the original problem before seeing analogies in Study 1, and those formulations can be attached to prompts. That 
Might the information of current user path enable LLMs to provide more personalized creative support for taking a deep dive or a “creative leap” \cite{chanSemanticallyFarInspirations2017,siangliulueProvidingTimelyExamples2015,noyQuantitativeStudyCreative2012}? We are curious how future work might fruitfully build on explorations by \cite{zhuBiologicallyInspiredDesign2023}, who used LLMs not just to generate analogies, but also generate explanations for potential mappings in an analogy, and generate potential concept ideas. We can imagine design patterns where an initial round of LLM-generated analogies helps to stimulate memory retrieval of diverse domains --- to assist with fixation \cite{linseyModalityRepresentationAnalogy2008, linseyRepresentingAnalogiesIncreasing2006} and surface similarity bias in analogical retrieval \cite{holyoakMentalLeapsAnalogy1996, gentnerAnalogicalRemindingGood1985, gickSchemaInductionAnalogical1983, gentnerRolesSimilarityTransfer1993} --- and then subsequent integration of LLM-assisted ``deep dives" via explanations of potential mappings, generating variations of ideas within a theme, and so on.

Another interesting direction may be to leverage LLM-generated analogies to improve the process of iteration from feedback \cite{chanImprovingCrowdInnovation2016,dowParallelPrototypingLeads2010,chanBestDesignIdeas2015}. Studies of analogies in creative work show that they are useful for more than concept generation and problem reformulation: analogies are frequently used to assist with explaining unfamiliar concepts \cite{dunbarHowScientistsThink1997,bearmanExplorationRealWorldAnalogical2002,christensenRelationshipAnalogicalDistance2007}. For example, the deployable space array technology is more explainable with an origami-folding analogy \cite{bolanosPreliminaryApproachSelect2022}. We wonder how AI-generated analogies could be used by creative workers not only to inspire themselves, but also to assist them in conveying design concepts to others by connecting to their domain knowledge. %Analogies can help the audience to transfer their previous knowledge and understanding to a new domain which they are unfamiliar with. 

\begin{acks}
This research was supported in part by ONR N000142012506.
\end{acks}

\bibliographystyle{ACM-Reference-Format}
\bibliography{sample-base}

% anonymity check
% \begin{acks}

% \end{acks}

\onecolumn

\newpage

\appendix

\textbf{APPENDIX}

% \vspace{-4.8mm}

\section{Screenshots of User Interface for Study 2}
\label{appendix:a}

\renewcommand{\thefigure}{A\arabic{figure}}

\setcounter{figure}{0}

\begin{figure*}[htp]
\centering \includegraphics[width=0.9\textwidth]{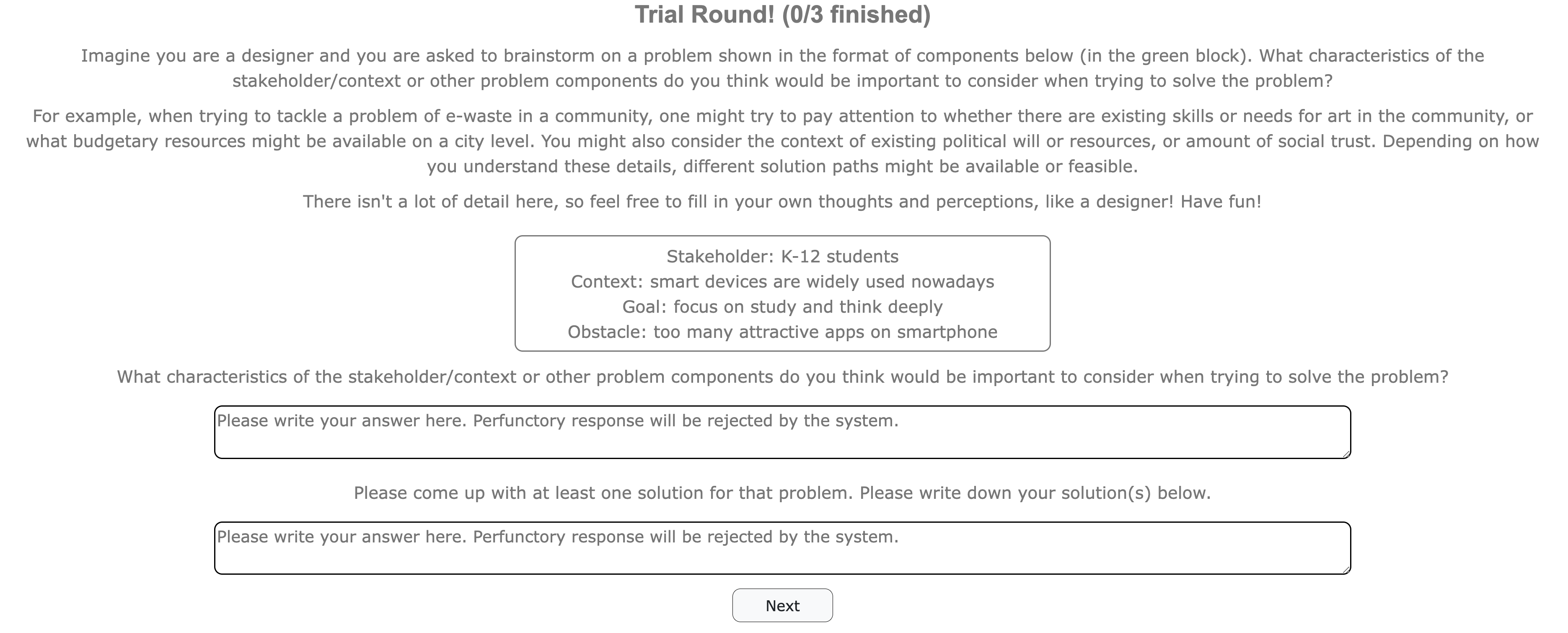}
\caption{Screenshot of the initial formulation phase task interface and instructions.}
\label{figa1}
\Description{TODO}
\end{figure*}

\begin{figure*}[htp]
\centering \includegraphics[width=0.9\textwidth]{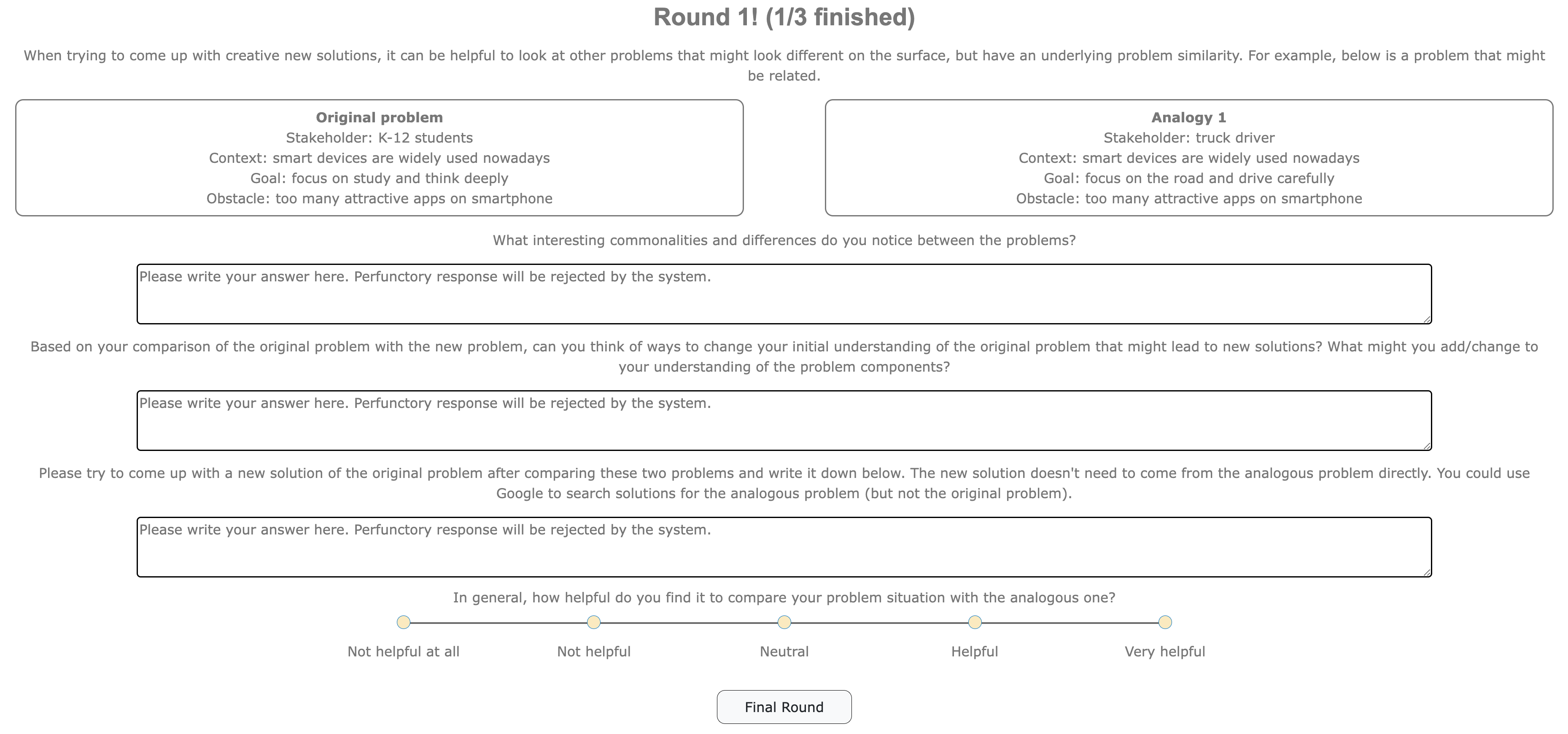}
\caption{Screenshot of the analogical reformulation and ideation phase task interface and instructions.}
\label{figa2}
\Description{TODO}
\end{figure*}

\end{document}